\documentclass[11pt, article, twocolumn]{article}
\usepackage[margin=0.8in]{geometry}
\usepackage{graphicx}

\usepackage{graphicx,soul}
\usepackage{amsmath}
\usepackage{bm}
\usepackage{epstopdf}
\usepackage{epsfig}
\usepackage{subcaption}
\usepackage{caption}
\usepackage{float}
\usepackage{hyperref}
\usepackage{xcolor}
\usepackage{soul}

\title{\textbf{{\Large Learning sparsity in reservoir computing through a novel bio-inspired algorithm  }}\vspace{-0.3cm}}
\author{Luca Manneschi$^1$ \\ {\small manneschi1@sheffield.ac.uk} \and Andrew C. Lin$^2$ \\ {\small andrew.lin@sheffield.ac.uk} \and Eleni Vasilaki$^1$ \\ {\small e.vasilaki@sheffield.ac.uk}
   }

\date{%
    $^1$Department of Computer Science, The University of Sheffield\\%
    $^2$Department of Biomedical Science, The University of Sheffield\\[2ex]%
    \today
    \vspace{-3ex}
}

\begin{document}

\maketitle


The mushroom body is the key network for the representation of learned olfactory stimuli in Drosophila and insects. The sparse activity of Kenyon cells, the principal  
neurons in the mushroom body, plays a key role in the learned classification of different odours. In the specific case of the fruit fly, the sparseness of the network is enforced by an inhibitory feedback neuron called APL, and by an intrinsic high firing threshold of the Kenyon cells. In this work we took inspiration from the fruit fly brain to formulate a novel machine learning algorithm that is able to optimize the sparsity level of a reservoir by changing the firing thresholds of the nodes.  The sparsity is only applied on the readout layer so as not to change the timescales of the reservoir and to allow the derivation of a one-layer update rule for the firing thresholds. The proposed algorithm is a combination of learning a neuron-specific sparsity threshold via gradient descent and a global sparsity threshold via a Markov chain Monte Carlo method. The proposed model outperforms the standard gradient descent, which is limited to the ``readout''weights of the reservoir, on two example tasks. It demonstrates how the learnt sparse representation can lead to better classification performance, memorization ability and convergence time.

\section{Introduction}

Sparsity is a well known concept in neuroscience, observed from the high selectivity of the neurons, ranging from the sensory cortex of mammalian brain \cite{rolls1995sparseness} to the Kenyon cells (KCs) in the mushroom body \cite{bhandawat2007sensory}. In particular, this work is inspired from the low coding level that is observed in the KCs of Drosophila, where the high intrinsic thresholds of KCs permits such neurons to sparsely and selectively encode external stimuli \cite{lin2014sparse}. Analogously, the model proposed exploits the concept of learnable thresholds to optimize the level of sparsity inside the network. 
The learning is performed by optimization of a distance measure between the output of the neural network and the desired outcome without exploiting any normalization term. The novelty of the proposed approach lies on the fact that a sparsity level is reached due to the presence of firing thresholds, rather than to regularization \cite{huang2011learning} \cite{hastie2015statistical} \cite{candes2008enhancing}.
From the machine learning perspective, adopting sparse representations can lead to more interpretable model \cite{hastie2015statistical}, to a reduced computational cost \cite{NIPS2016_6504}, and can help solve overfitting problems \cite{srivastava2014dropout}. 
In this regard, the work in \cite{NIPS2016_6504} demonstrated how structured sparsity can have benefits in term of computational speed and accuracy in a convolutional neural network. Rasmussen et al. \cite{rasmussen2012model} showed how the choice of regularization parameters of the model can impact the interpretability and the reproducibility of a classifier of neuroimaging data, and showed the existence of a trade-off between pure classification accuracy and reproducibility. The Dropout technique \cite{srivastava2014dropout}, which selects random subset of units in a neural network during training, can prevent overfitting by diminishing the codependence of the units in deep neural networks.
\newline
The network under consideration in this work is a reservoir of leaky integrators \cite{jaeger2007optimization}. The connectivity between the nodes is represented through a random sparse fixed adjacency matrix that enables the associated dynamical system to exhibit a multitude of characteristic timescales. This complex connectivity is consistent with experimental reports of chemical \cite{takemura2017connectome} and electrical \cite{liu2016gap} synapses between Kenyon cells in Drosophila, although the physiological function of KC-KC synapses has not yet been discovered.

\section{Methods}

The reservoir under consideration is a network of leaky integrators described by the following equation

\begin{equation}
\textbf{V}(t+1)=(1-\alpha)\textbf{V}(t)+\alpha f\big[W_{in}\textbf{s}+\rho W\textbf{V}(t)\big]
\label{V}
\end{equation}

where $\alpha=\frac{\delta t}{\tau}$ defines the temporal scale of the neuron and $\textbf{V}(t)$ is the activity vector of the integrators \footnote{It is called $\textbf{V}$ to resemble the voltage of a neuron.}. The activation function $f$ chosen is a rectified linear unit. $W_{in}$ is the input adjacency matrix, $W$ is the fixed sparse random matrix that describes the recurrency of the reservoir, and $\textbf{s}$ is the signal. The rescaling factor $\rho$ is chosen in order to constrain the eigenvalues of the associated dynamic system inside the imaginary plane and guarantee the Echo State property. The number of connections of the input matrix $W_{in}$ from the input neurons to the reservoir follow a lognormal distribution where each node in the reservoir is connected to {\it six} input nodes on average. Furthermore, a specific weight of a postsynaptic neuron $i$ is inversely proportional to the number of connections to such neuron. This choice of $W_{in}$ is inspired by experimental evidence \cite{caron2013random}.
However, other forms of $W_{in}$ are possible and the results of this work do not depend considerably on the specific form of the distribution of the input connections. Nevertheless, the magnitude of the weights of $W_{in}$ plays an important role and must be chosen appropriately \cite{jaeger2007optimization}. \newline \newline   
In contrast to previous models \cite{jaeger2001echo} \cite{jaeger2002tutorial} \cite{jaeger2007optimization} that define the output of the neural network through a read-out of the $\textbf{V}$ vector, we introduced another variable $\textbf{x}(t)$, defined as follows

\begin{equation}
\textbf{x}(t)=relu\big[ \textbf{V}(t)-\bm{\theta} \big]
\label{x}
\end{equation}
where $relu$ stands for rectified linear unit, and $\bm{\theta}$ is a vector of thresholds that enables $\textbf{x}$ to be sparse. Thus, the `measurable' variable $x_i(t)$ is zero if the `hidden' variable $V_i(t)$ is lower than the corresponding threshold $\theta_i$.     
The training procedure minimizes a measure of the distance $E(t)$ between the output $y(t)=W_{out}\textbf{x}(t)$ of the neural network and the desired value $y_{true}(t)$. Mathematically, the cost function is
 
\begin{equation}
E=\sum_j \Big[y^{true}_j-\sum_i W^{out}_{ji}relu\Big(\textbf{V}_i(t)-\theta_i\Big)\Big]^2
\label{E}
\end{equation}

in which $W_{out}$ and the thresholds $\bm{\theta}$ are the learnable parameters.
Thanks to eq.\ref{x} and the introduction of the sparse measurable variable $\textbf{x}$ it is possible to change the thresholds without affecting the temporal timescales of the reservoir. This specific formulation allow us to compute the gradient of $E$ with respect to $\theta$ without incurring  backpropagation through time and to preserve the idea behind reservoir computing as a fixed, dynamically rich, representation. 
The rest of this methodological part is organized as follows: subsection $2.1$ describes the tasks analysed, in subsection $2.2$ we consider two possible training procedure to learn the thresholds values, and subsection $2.3$ describes the best performing algorithm.

\subsection{Tasks}
We considered two classification tasks where the model must make a decision after a prefixed time interval $\Delta t$. To make the task more challenging and more biologically plausible, the model selects a class \footnote{Or an action in the case of Reinforcement Learning that will be analysed later.} with probability that corresponds to a softmax function applied on the output layer. It follows that choosing the class that corresponds to the highest output would only make the task easier for all the models that will be analysed, and that the relative differences among the various algorithms and our conclusions would not change. \newline 

\begin{figure*}[h!]
 \centering
    \makebox[\textwidth][c]{\includegraphics[width=1.\textwidth]{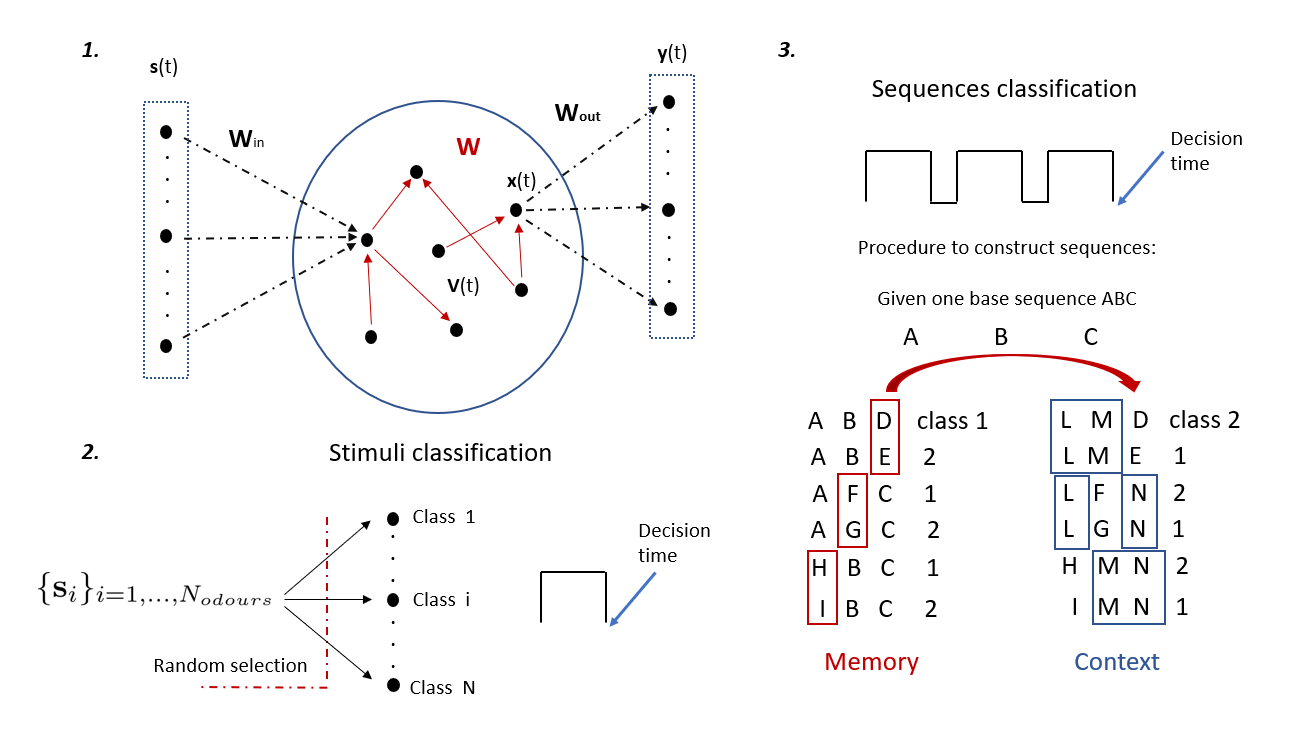}}
    \caption{Scheme of the network and of the tasks considered. \textbf{1.} The network is composed by a reservoir of integrators. The activities of the nodes is divided into two variables: a hidden variable $\textbf{V}(t)$ containing all the complex dynamic  of the reservoir, and a measurable variable $\textbf{x}(t)$ containing the thresholds values and defining a sparse output representations. \textbf{2.} Illustration of the static task considered. The network has to classify experimental stimuli perturbed with white noise and randomly associated to different classes. The model has to classify at the end of the presentation of the external stimulus. \textbf{3.} Scheme of the classification task and of the procedure adopted to define sequences from single stimuli (\textit{Methods} for more details). As before, the network is asked to classify at the end of the successions which are assigned to random classes.   }
    
  \label{Figure1}
\end{figure*}

In the first paradigm, the external input s(t) is derived from the simulated response of 24 projection neurons (PNs, second-order neurons in the fly olfactory system) to 110 different odors, based on physiological recordings of olfactory receptor neurons (ORNs) and known characteristics of the ORN-PN synapse \cite{hallem2006coding} \cite{olsen2010divisive}. This simulated activity, which we call $s^{HO}$ (HO for Hallem-Olsen), has previously been used in computational analyses of fly olfaction \cite{luo2010generating} \cite{parnas2013odor} \cite{krishnamurthy2017disorder}. 
If $\textbf{s}^{HO}$ is the $N_{In}$ dimensional vector describing the activities of the input neurons, the i-th dimension of the input is $s_i(t)=s^{HO}_{i}+\sigma\xi(t)s^{HO}_i$, where $\xi(t)$ is a Gaussian distributed random variable with zero mean and unitary variance. Thus, the temporal dependence of the external stimulus is due to the presence of noise only and the stimulus, without carrying any relevant temporal information, is practically static\footnote{Indeed, a simple network of unconnected integrators can solve this first experiment successfully as long as the characteristic times of the nodes is big enough to smooth the fluctuation of the signal.}. Each stimulus $\textbf{s}_j(t)$ is then associated to a random chosen class, which is the desired outcome of the classification. Given the random nature of the pairing between stimuli and corresponding correct outputs, the model cannot exploit correlations among different signals and the performance achieved is a measure of pure memorization ability. A scheme of the task is shown in fig.\ref{Figure1} \newline \newline
In the second paradigm considered we evaluated the performance of the models in classifying sequences of three successive stimuli. The procedure for building different sequences is described in the third panel of fig.\ref{Figure1}. Given a base sequence of randomly selected stimuli $ABC$ \footnote{The improbable choice of the stimuli $ABC$ is for illustrative purposes.}, we substituted the last signal $C$ with $N_{class}$ random stimuli (if $N_{class}=2$, $D$ and $E$ for instance) and associated each new sequence to a random different class (following the considered example, $ABD$ to class one and $ABE$ to class two). From now on, we will call ``perturbations" the new substituted elements of the sequence and "context" the remaining elements. In the case of $AFC$, which is derived from the base $ABC$, $F$ is the perturbation while $AC$ represents the context.  Then, the same procedure is applied to the previous elements $B$ and $A$ of $ABC$ as illustrated in the left column of the third panel of fig.\ref{Figure1}. The purpose of this procedure of defining sequences through perturbations of single elements is to test the temporal memory, or echo, of the reservoir. For a given sequence $\textbf{S}_i(t)$, the network has to remember the association between a perturbation and the desired output class. However, if perturbations are not repeated, the model does not have to take into account the relationships among various elements, but the memorization of the perturbation is sufficient to obtain a correct classification. In order to make the task more challenging, we changed the context of a perturbed sequence $\textbf{S}_i(t)$ by replacing the two context elements with new random stimuli. The new succession is then associated to a different class than the original $\textbf{S}_i(t)$. For instance, if $ABD$ is the sequence obtained by perturbing the base $ABC$, a contextual change can be obtained by replacing $AB$ with $LM$ and by defining the new succession $LMD$.
Contextual variations are finally illustrated in the right column of the the third panel of fig.\ref{Figure1} The overall procedure defines a systematic way to test the temporal and contextual memory capacity of the network while preserving the random association between desired class and successions of elements. Apart from possible similarities among sequences of the same class due to a poor statistics, there are no correlations between inputs that could help the classification procedure, and each sequence has to be classified independently. Thus, given $N_{base}$ successions, this methodology creates a total number of sequences that equals to $3N_{class}^{2}N_{base}$, where the factor $3$ is due to the number of elements in a succession, a factor $N_{class}$ to perturbations, and a factor $N_{class}$ to contextual changes.  

\subsection{Algorithms}

The objective of this section is to review two possible algorithms to optimize the vector of thresholds $\bm{\theta}$ of equation \ref{x} capable of learning sparse representations of input stimuli and to introduce a benchmark model. The first one is based on the derivation of the gradient of the error function with respect to the individual neuronal thresholds while the second one is the well known Metropolis algorithm. By optimizing a sparsity level, we envisage that the desired learning rule would also separate the representation of stimuli belonging to different classes in order to facilitate the classification, as non-specific neurons, i.e. neurons that fire to all stimuli, are likely unhelpful for the learning process. The decrease of the overlap between antagonist representations will be quantified through a specificity measure that is introduced in the next section.

\begin{flushleft}
\textit{\textbf{Benchmark ($\mathbf{GD}_{W}$)}}
\end{flushleft}

The benchmark model exploits gradient descent on the output weights $W^{out}$ without the introduction of the thresholds and the additive layer $\textbf{x}(t)$. The output activity corresponds to a readout of the $\textbf{V}$ variable (eq.\ref{V}). \newline
Also all the other algorithms analysed will learn $W^{out}$ through gradient descent, but their output activity will be a readout of the $\textbf{x}$ (eq.\ref{x}) variable and they will have the additional complexity of learning the thresholds $\bm{\theta}$.

\begin{flushleft}
\textit{\textbf{Gradient Descent on $\bm{\theta}$ ($\mathbf{GD}_{\theta}$)}}
\end{flushleft}
In addition to learning the 'output' weights of the reservoir with gradient descent, we choose to learn the local firing thresholds. The derivative of the cost function $E$ with respect the $\theta_i$ leads to

\begin{flalign}
\nonumber
\Delta \theta_i =-\eta \dfrac{\partial E}{\partial \theta_i}= \\\nonumber
 \eta \sum_{j=1}^{N_{class}} \big[ y^{true}_j-y_j(t) \big] W^{out}_{ji} \dfrac{\partial relu\big[V_i(t)-\theta_i\big]}{\partial \theta_i}=\\\nonumber
=-\eta\sum_{j=1}^{N_{class}} \big[ y^{true}_j-y_j(t) \big] W^{out}_{ji}H\big(V_i(t)-\theta_i\big)=\\
=-\eta\sum_{j=1}^{N_{class}} \big[ y^{true}_j-y_j(t) \big] W^{out}_{ji}H\big(x_i(t)\big) \label{theta_gradient}
\end{flalign}

where $\eta$ is the learning rate and $H$ is the Heaviside function. While this is a supervised scenario, we are adopting a Reinforcement Learning terminology: eq.\ref{theta_gradient} can be viewed as the sum of prediction errors where each reinforcement signal belongs to a different class and is modulated by the corresponding output synapses. 
In order to interpret eq.\ref{theta_gradient}, let us consider the case in which $N_{class}=2$ so that we have a qualitative idea of what the meaning of eq.\ref{theta_gradient} is. In this specific scenario eq.\ref{theta_gradient} becomes: 

\begin{flalign}
\nonumber
\scriptstyle \Delta \theta_i =-\eta \dfrac{\partial E}{\partial \theta_i}=\\\nonumber
\scriptstyle =-\eta\Big\{\big[ y^{true}_1-y_1(t) \big]W^{out}_{1i}+[ y^{true}_2-y_2(t) \big]W^{out}_{2i}\Big\} H\big(x_i(t)\big)=\\\nonumber
\scriptstyle =-\eta\Big\{\big[ \tilde{y}-y_1(t) \big]W^{out}_{1i}-y_2(t)W^{out}_{2i}\Big\} H\big(x_i(t)\big)
\end{flalign}
where we consider that desired output is a positive quantity $\tilde{y}$, usually set as one in a classification task for the correct class and zero otherwise. Let us consider the extreme assumption where $y^{true}_1-y_1(t) \approx y^{true}_2-y_2(t)$ and a first learning phase where the relations $y^{true}_1-y_1(t)>0$ and $y^{true}_2-y_2(t)<0$ can hold by construction. In this ideal case the learning rule is driven by the difference $W^{out}_{1i}-W^{out}_{2i}$, that leads to a decrease of the i-th firing threshold when $W^{out}_{1i}>W^{out}_{2i}$ and the i-th specific node is helping to achieve the right classification, and an increase of the threshold value when $W^{out}_{1i}<W^{out}_{2i}$ and the node is contributing to reach the wrong output. 
The general case where $y^{true}_1-y_1(t) \neq y^{true}_2-y_2(t)$ is now understandable by considering that the local factor $W$ are modulated through a feedback signal that gives the priority to nodes that are far from the desired output. \newline
The problem with this gradient based rule is that it changes the activation of the nodes unidirectionally. While eq.\ref{theta_gradient} deactivates nodes that are not useful for the classification task, the learning rule cannot reactivate neurons that were silent because the gradient of a non active node is zero by definition. 
Furthermore, we will later see that the mean of the distribution of thresholds found by this algorithm is suboptimal. \footnote{This last consideration will be supported by the results of fig.\ref{Figure3}.}
The limitations of a gradient based approach led us to consider the following algorithm.
 
\begin{flushleft}
\textit{\textbf{Metropolis algorithm ($\mathbf{Metropolis}_{\theta}$)}}
\end{flushleft}
While the previous algorithm optimizes a separate threshold for each neuron, the algorithm analysed in this section exploits a global threshold for the whole network ($\bm{\theta}=\theta$). Indeed, a desired sparsity level is reachable with one single value of $\theta$ only, and preliminary simulations have demonstrated how adopting diverse fixed values of the global threshold can lead to different performance. We want to analyse the consequences of optimizing such parameter with an algorithm that, instead of using the derivative of the cost function, performs a stochastic search by randomly perturbing the value of $\theta$ and accepts or declines the new value with some probability. In the proposed implementation such a probability is given by the Metropolis algorithm \cite{kuczera1998monte} where the energy of the system corresponds to the cost function $E$. The procedure adopted is summarized in the following steps:
\begin{itemize}
\item[(i)] \textit{Starting from a value of $\theta^{-}$, propose a new threshold value $\theta^{+}$, where} \begin{center} $\theta^{+}=\theta^{-}+\sigma_M N(0,1)$ \end{center}  
\item[(ii)] \textit{Repeat for $M$ steps:} \newline \newline
\hspace*{3ex} \textit{Compute the cost $E^{\pm}=\sum_{j}\big(y^{true}_j-y^{\pm}_j\big)^2$ \hspace*{3ex} for the two thresholds values $\theta^{\pm}$} \newline \newline
\hspace*{3ex} \textit{Update the output weights $W^{\pm}$ through
\hspace*{3ex} gradient descent on $E^{\pm}$} \newline \newline
\hspace*{3ex} \textit{Compute an average $\mathcal{E^{\pm}}$ of $E^{\pm}$ }
\begin{center} $\mathcal{E^{\pm}}=(1-\alpha_M)\mathcal{E^{\pm}}+\alpha_M E^{\pm}$  \end{center} 
\item[(iii)] \textit{Accept the network corresponding to $\theta^{+}$ with probability  \begin{center} $p=min\Big\{1,\exp \big(-\beta(\mathcal{E}^{+}-\mathcal{E}^{-} )\big)\Big\}$ \end{center} } 
\end{itemize} 

Practically, the algorithm proposes a new reservoir with a global threshold $\theta^{+}$, changes the output weights through standard gradient descent, and accepts the new network by applying the Metropolis rule on a running exponential average of the cost function.
In the above scheme, $M$ is the number of steps used to compute $\mathcal{E}$ where the values of $\theta^{\pm}$ are fixed, and $\alpha_M \propto 1/M$ defines the memory of the running average . 
 
\subsection{Proposed algorithm, a unified approach} 
 
The proposed algorithm exploits a mixture of the Metropolis and the gradient descent updating rules to change the values of the thresholds. Each single node has a threshold
\begin{equation}
\theta_i=\theta_g+\tilde{\theta}_i
\end{equation}   
defined as the sum of a global factor $\theta_g$ and a local factor $\tilde{\theta}_i$. The proposed model optimizes the global part through stochastic perturbations and the local one via a gradient descent approach. 
The algorithm shares the same steps of the \textit{Metropolis} procedure explained above (section \textit{Metropolis}), with the following differences: the stochastic perturbation on step $(i)$ is applied to the global factor $\theta_g$, and 
the gradient descent on step $(ii)$ is also applied to the local thresholds (instead of the output weights only).    
For clarity, the scheme of the final algorithm is  
 
\begin{itemize}
\item[(i)] \textit{Starting from a value of $\theta_g^{-}$, propose a new threshold value $\theta_g^{+}$, where} \begin{center} $\theta_g^{+}=\theta_g^{-}+\sigma_M N(0,1)$ \end{center}  
\item[(ii)] \textit{Repeat for $M$ steps:} \newline \newline
\hspace*{3ex} \textit{Compute the cost $E^{\pm}=\sum_{j}\big(y^{true}_j-y^{\pm}_j\big)^2$ \hspace*{3ex} for the two thresholds values $\theta^{\pm}$} \newline \newline
\hspace*{3ex} \textit{Update the output weights $W^{\pm}$ and $\tilde{\theta}_i^{\pm}$ 
\hspace*{3ex} through gradient descent on $E^{\pm}$} \newline \newline
\hspace*{3ex} \textit{Compute an average $\mathcal{E^{\pm}}$ of $E^{\pm}$ }
\begin{center} $\mathcal{E^{\pm}}=(1-\alpha_M)\mathcal{E^{\pm}}+\alpha_M E^{\pm}$  \end{center} 
\item[(iii)] \textit{Accept the network corresponding to $\theta_g^{+}$ with probability  \begin{center} $p=min\Big\{1,\exp \big(-\beta(\mathcal{E}^{+}-\mathcal{E}^{-} )\big)\Big\}$ \end{center} } 
\end{itemize}  
 
Since the algorithm is now learning three sets of variables, $\theta_i$, $\theta_g$ and $W^{out}$, one important aspect to take into account is the fact that the effective learning rate depends on the sparsity level. In particular, changing the initial value of $\theta_g$ can strongly affect the effective learning rate. Because  the optimization of the weights and of the local thresholds depends on the step size and consequently on $\theta_g$, the initial value of $\theta_g$ can become important even if the stochastic variations of the \textit{Metropolis} are optimizing it.
In order to choose the starting condition, the algorithm exploits a small prelearning phase (about $10000$ steps) by applying the same steps described in the scheme above, but resetting $W_{out}$ and $\theta_g$ after each \textit{Metropolis} update to their initial values and trying different initial values of $\theta_g$. In such a way, we guarantee choosing a good starting condition. In practice, this procedure was done only one time for each task considered without trying to fine tune the initial $\theta_g$ for all the simulations performed.   
 
\subsection{Specificity}

Since a sparsity level is obtained through direct minimization of eq.\ref{E} by gradient descent and/or stochastic search, we expect that the optimized thresholds would decrease the overlap in the representations among stimuli corresponding to different classes. Indeed, this separation would make the classification task easier. Thus, a measure of specificity is formulated to quantify and to understand how the proposed learning can lead to better representations.
Let us consider two classes $j$ and $k$ and a neuron $i$. The node is specific to the class $i$ with respect to the other if it is more active when stimuli belonging to class $i$ are presented to the network. Generalizing this idea it is possible to build a tensor $spec_{ijk}$ defined as

\begin{equation}
spec_{ijk}=\frac{|N_{ij}-N_{ik}|}{N}
\label{specificity}
\end{equation} 

where $N_{ij}$ ($N_{ik}$) are the number of times the neuron $i$ was active after the presentation of a stimulus of class $j$ ($k$) and $N$ is the total number of  episodes taken into account to compute eq.\ref{specificity}.
There are also other possibilities analogous to eq. \footnote{$N$ could be substituted by $N_{ij}+N_{ik}$ and define a relative specificity that is rescaled by the number of times a neuron was active. However, a decrease of $N_{ij}+N_{ik}$ causes an undesirable increase in the specificity measure. Another choice is to use the sum of the activities of a neuron for all the stimuli in class $\sum_j x_{ij}$ instead of $N_{ij}$, but this leads to comparable results.}, but we considered the above equation the most easily interpretable.
Given $spec_{ijk}$ it is possible to compute a measure of specificity for each single neuron as

\begin{equation}
Sp_{i}= \frac{1}{(N_{class}-1)!}\sum_j \sum_{k>j} spec_{ijk}
\end{equation}  
where we considered only the upper triangular part of $spec_{ijk}$ because of the symmetry of the latter tensor.
 
\begin{center}
\begin{tabular}{ |p{3cm}||p{3cm}|  }
\hline
  Parameters & Values, $stimuli/sequence$ task \\
 \hline
 \hline
 \multicolumn{2}{|c|}{Tasks} \\
 \hline
 $N_{class}$   & $2$   \\
 $ \sigma$  &   $0.3/0.2$  \\
 $\Delta_t$ & $0.5/0.3 s $  \\
 \hline
 \multicolumn{2}{|c|}{Network} \\
 \hline
 $\alpha $    & $0.025/0.1$ \\
 $\rho$ &   $0.8/0.95$  \\
 $N$ & $1000$ \\
 $N_{In}$ & $24$ \\ 
 \hline
 \multicolumn{2}{|c|}{Model} \\
 \hline
 $\eta_{W}$ & $0.0018$  \\
 $\eta_{\theta}$ & $0.00018$ \\
 $N_{batch}$ & $1/10$ \\
 $\sigma_M$ & $0.05$ \\
 $N_M$    & $100$ \\
 $\beta$ & $4$ \\
 \hline
\end{tabular}
\end{center}

\section{Results}  

We will now compare the four algorithms: gradient descent on the weights ($GD_{W}$), gradient descent on the weights and thresholds ($GD_{\theta}$), the Metropolis algorithm $Metropolis_{\theta}$ and the composite model on the two tasks (section \textit{Tasks}).

\begin{flushleft}
\textit{\textbf{Task 1, static input}}
\end{flushleft}

\begin{figure*}[h!]
 \centering
    \makebox[\textwidth][c]{\includegraphics[width=1.2\textwidth]{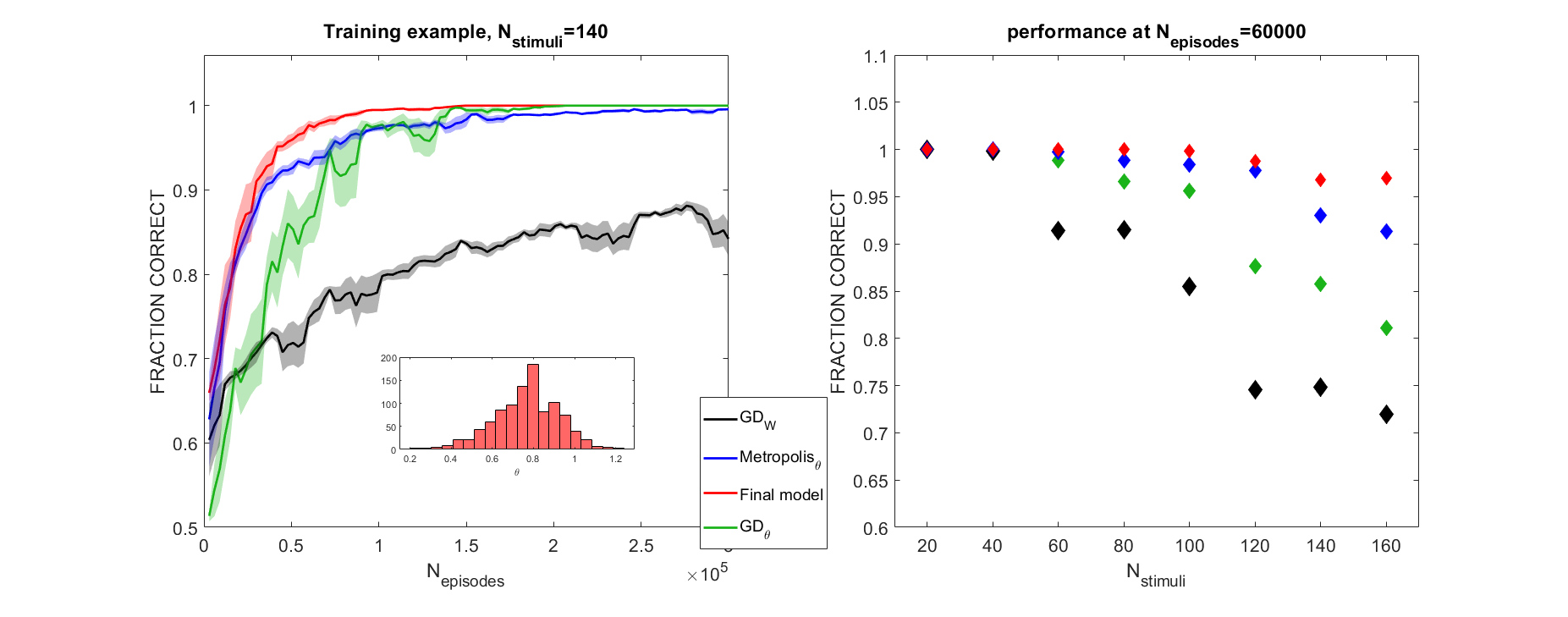}}
    \caption{\small{Task 1, performance. \textbf{Left}. Performance of the model during training. The \textit{Composed model} has the highest speed of convergence, while the model without thresholds and sparse activity has lowest accuracy. \textbf{Right} Fraction of correct classifications after $60000$ episodes. The difference between the algorithms increases with the difficulty of the task.  }}
    
  \label{Figure2}
\end{figure*}

In this scenario the model has to classify, after a fixed time interval of $\Delta t=0.5s$, a noisy $24$ dimensional stimulus. Each input is defined as described in the methodological section and it is randomly assigned to a class in order to test the pure memorization ability of the network. 
The performance is shown in fig.\ref{Figure2}, where the left panel reports a training example of a classification of $140$ stimuli. The parameters defining the task and the hyperparameters of the composed algorithm are reported in the table above.\newline

\begin{figure*}[h!]
 \centering
    \makebox[\textwidth][c]{\includegraphics[width=1.2\textwidth]{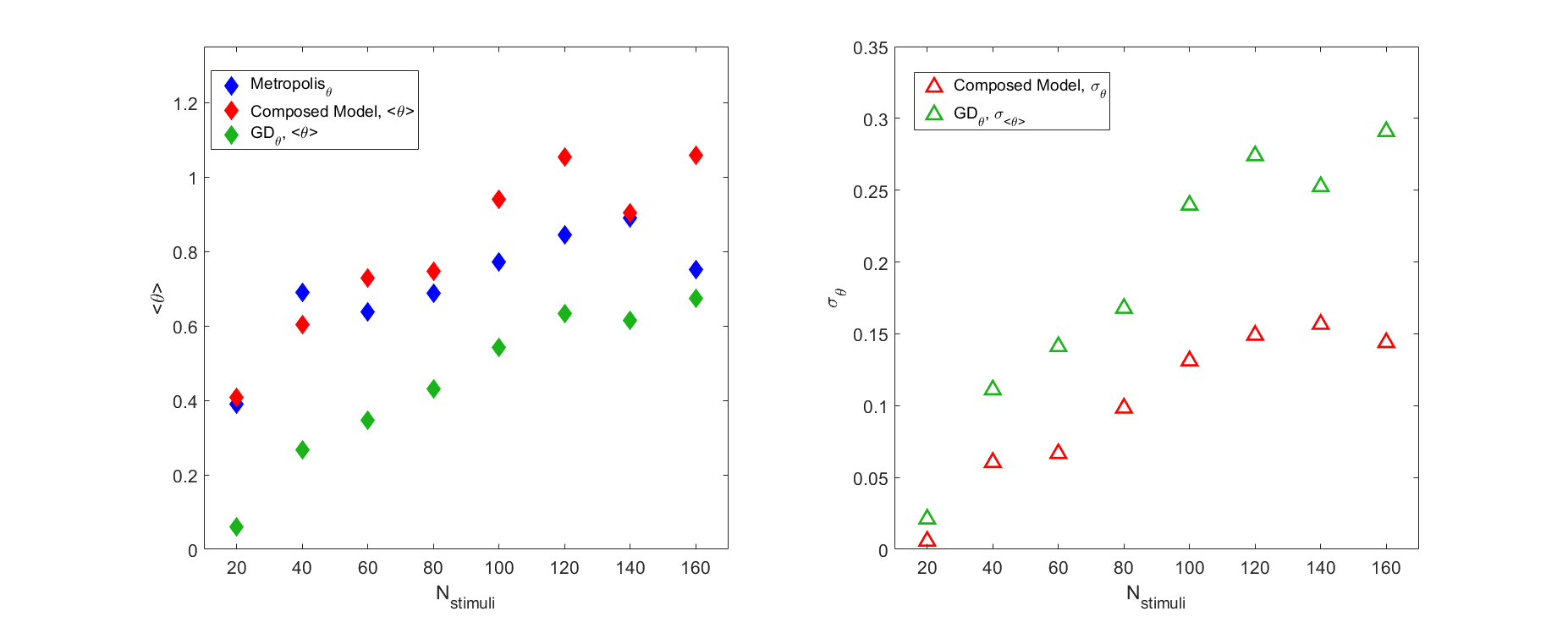}}
    \caption{\small{Task 1, Mean and variance of the optimized $\theta$ distribution. \textbf{Left} The average of the distribution shows an upward trend and the sparseness in the network rises with the number of input stimuli. \textbf{Right} The need to differentiate the values of the thresholds is reflected in the $\sigma$ of the optimized distribution and it increases as the task becomes more demanding. }}
    
  \label{Figure3}
\end{figure*}

The black line refers to the standard model called $GD_{W}$ \footnote{for gradient descent on the output weights} , where the threshold $\theta$ and the $\textbf{x}$ variable are not introduced; the output of the neural network is a readout of $\textbf{V}$. Since the algorithms proposed are more complex and can exploit an exponential running average of the cost function to update the global parameter, we optimised the batch size of $GD_{W}$ to improve its performance and to make the comparison to the disadvantage of the proposed algorithm. The batch size chosen for $GD_{W}$ is $N_{batch}=100$. It is clear how the three models that exploit the additional complexity due to the thresholds outperform the standard model in terms of convergence speed and accuracy. The performance is also reported in the right panel after $60000$ episodes \footnote{One episode corresponds to a single presentation of a stimulus that ends when the network makes a decision} as the number of stimuli to be classified changes and the task becomes more demanding. The difference between the models become more evident as the difficulty increases, and the composite model robustly reports the best classification accuracy. 

After learning and for the case where $N_{stimuli}=140$,  the average percentage of active nodes is about fifty percent for the composed algorithm and the Metropolis, while it is about seventy percent for $GD_{\theta}$. This difference in the sparsity levels found by the three algorithms is visible in the left panel of fig.\ref{Figure3}, where the variance and the mean of the optimized distribution of $\theta$ are reported for all the cases. The \textit{Metropolis} and the composed model report higher values of the mean of the thresholds distribution (blue and red diamonds) in comparison to $GD_{\theta}$, which leads to a corresponding higher coding level. However, the increasing trend of $<\theta>$  for all the learning rules demonstrate how sparsity is needed when the memorization ability required by the network is high. The importance of a local threshold is reflected on the spread of the distribution, reported as $\sigma_{\theta}$. Indeed, as the performance between the \textit{Metropolis} and the composed model diverge going from left to right (right panel of fig.\ref{Figure2}), the variance of the optimized distribution raises (fig.\ref{Figure3}). \newline
Thus, even if learning a single global parameter leads to remarkable improvements compared to $GD_{W}$, learning a distribution of $\theta$ and optimizing local parameters can become relevant when the model has to memorize a large number of inputs. The surprising results obtained by optimizing a global factor through stochastic changes are due to the static nature of the signal used for this task, and the importance of the local thresholds will become dominant for sequence classification.

\begin{flushleft}
\textit{\textbf{Task 2, sequence classification}}
\end{flushleft}
Each input corresponds to a noisy succession of three stimuli that is defined as described in \textit{Methods}. A single element of the succession lasts for $0.1s$ and the network is asked to classify after $\Delta t=0.3s$. Sequences are randomly associated to classes as in the previous task, but in this case the model has to temporally remember the past of the signal and to take into account relationships among elements of the sequence to classify correctly. For this specific case, we used a batch size of $N_{batch}=10$ for $GD_{\theta}$, \textit{Composed model} and \textit{Metropolis}, while a batch size of $100$ for the standard algorithm $GD_{W}$.
The batch size for $GD_{W}$ was chosen to optimize its performance in terms of speed and accuracy. 
 
\begin{figure*}[h!]
 \centering
    \makebox[\textwidth][c]{\includegraphics[width=1.1\textwidth]{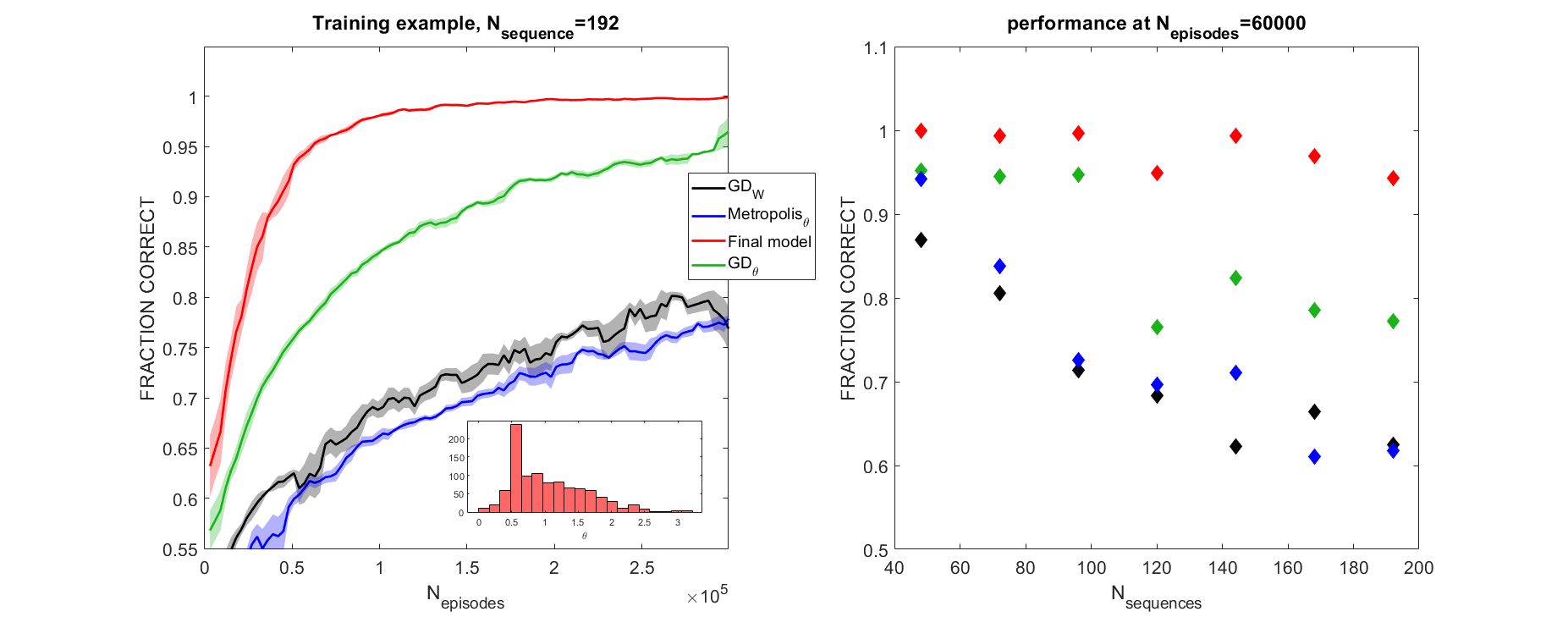}}
    \caption{\small{Task 1, performance. \textbf{Left}. Performance of the model during training. The performance of the composed model are the highest. The most remarkable difference in comparison to the results of \textit{Task 1} is the low accuracy reported by the \textit{Metropolis}, which is the model that optimizes a global threshold. Since  the task considered is dynamic, this result is expected (see text). }}
    
  \label{Figure4}
\end{figure*} 
 
 \begin{figure*}[h!]
 \centering
    \makebox[\textwidth][c]{\includegraphics[width=1.1\textwidth]{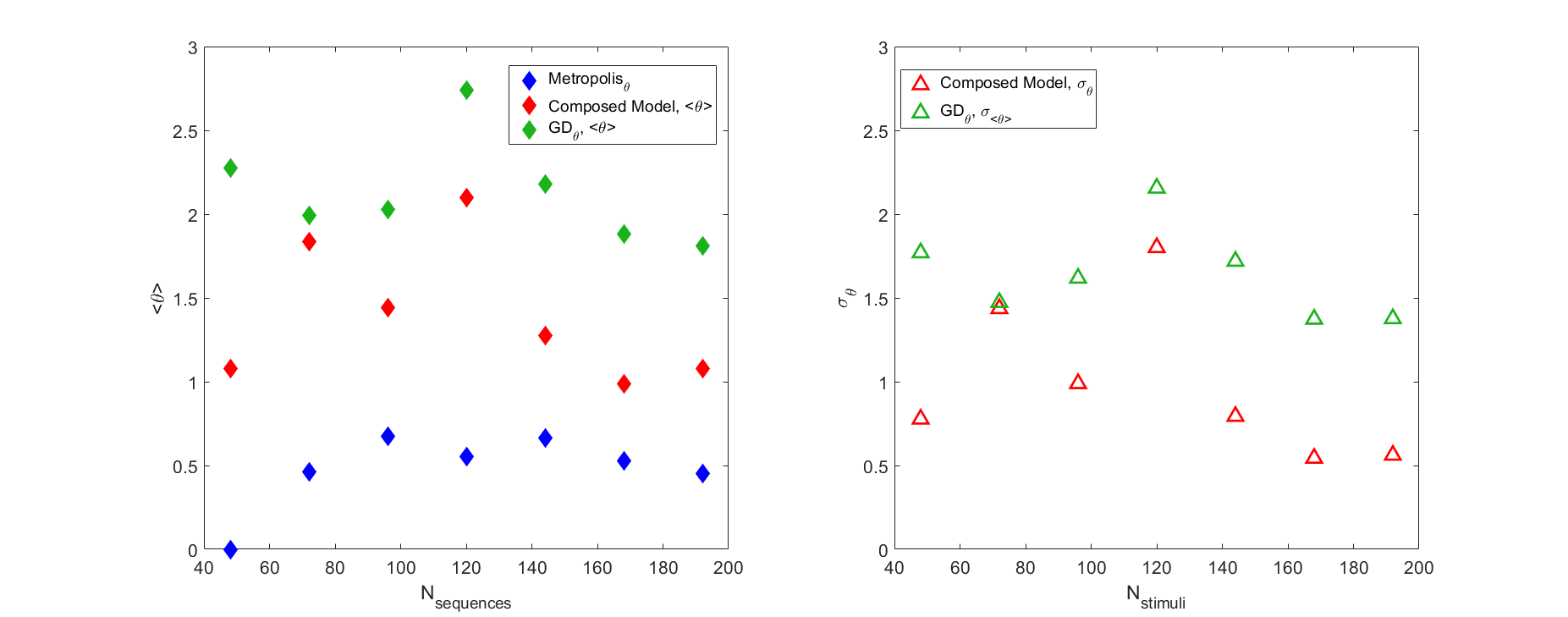}}
    \caption{\small{Task 2, Mean and variance of the optimized $\bm{\theta}$ distribution. Even if there is no evident trend in the average or the variance of the distribution, the results showed in fig.\ref{Figure3} are robust with respect to the variations of the values showed. $GD_{\theta}$ reports the highest mean, which is suboptimal if the performance of fig.\ref{Figure2} are considered. Thus, the presence of the global threshold in the \textit{Composed model} helps the model to reach a better $\bm{\theta$} distribution and a corresponding sparsity level.}}
    
  \label{Figure5}
\end{figure*}

 In this case the \textit{Metropolis} model is not able to concretely boost the performance of the $GD_{W}$ model (fig.\ref{Figure4}). Considering the dynamic nature of the stimuli and the consequent difficulty of imposing a shared fixed threshold on the nodes, this result is expected. Indeed, low activities of the nodes can decode important features of the dynamic signal. As a consequence, the spread of the distribution of $\theta$ is higher than in the static task. The difference between the performance of $GD_{\theta}$ and the composed algorithm can be better understood by considering the results reported in the left panel of fig.\ref{Figure5}. The $GD_{\theta}$ model reports a too high average value of the $\theta$ distribution. Also from the results of fig.\ref{Figure3} it was clear that the gradient descent based model was not able to find an optimal mean value of $\theta$, and the results found in this task confirm the need for a global threshold. We need also to specify that the pretraining procedure of the composed model described above is critical in this case, and that different starting values of $\theta_g$ can lead to totally different results even for the composed model. This is because the gradient descent shapes the distribution of $\theta_i$ very quickly. In other words, the rapid optimization of $\theta_i$ can trap the algorithm on a local minimum from which the slower stochastic perturbation of $\theta_g$ can hardly escape. The problem is fixed with a prelearning phase through the application of the \textit{Metropolis} algorithm, but genetic algorithms or an adaptive learning rate can also solve this complication.
However, the results confirm how an optimized sparsity level achievable due to the concept of firing thresholds can dramatically improve performance.

\begin{figure*}[htb!]
 \centering
    \makebox[\textwidth][c]{\includegraphics[width=0.9\textwidth]{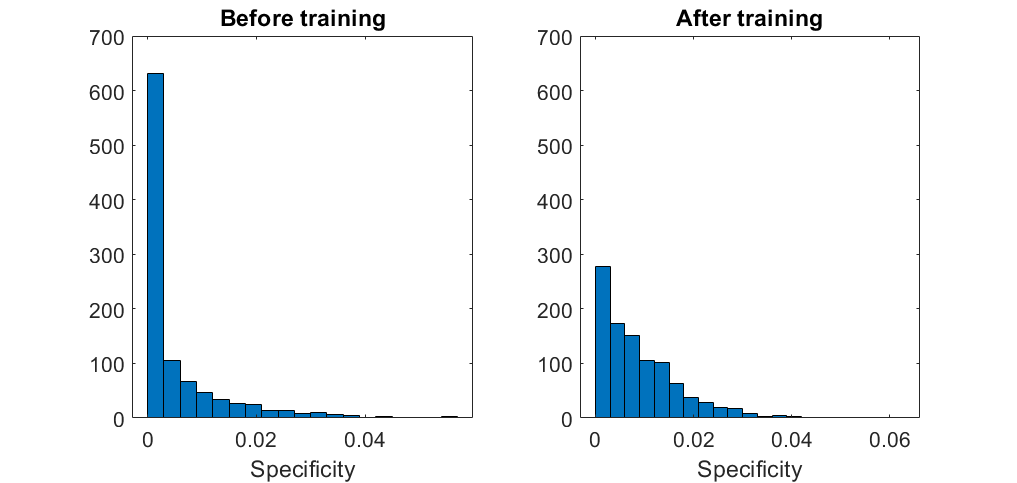}}
    \caption{\small{Change in the level of specificity after training. \textbf{Left}. Distribution of $Sp_{i}$ before learning. \textbf{Right}. Distribution of $Sp_{i}$ after learning. }}
    
  \label{Figure6}
\end{figure*}

Fig.\ref{Figure6} shows an example of specificity change before and after learning, demonstrating how the proposed algorithm is able to decrease the overlap among representations corresponding to different classes.

\begin{figure*}[h!]
 \centering
    \makebox[\textwidth][c]{\includegraphics[width=1.2\textwidth]{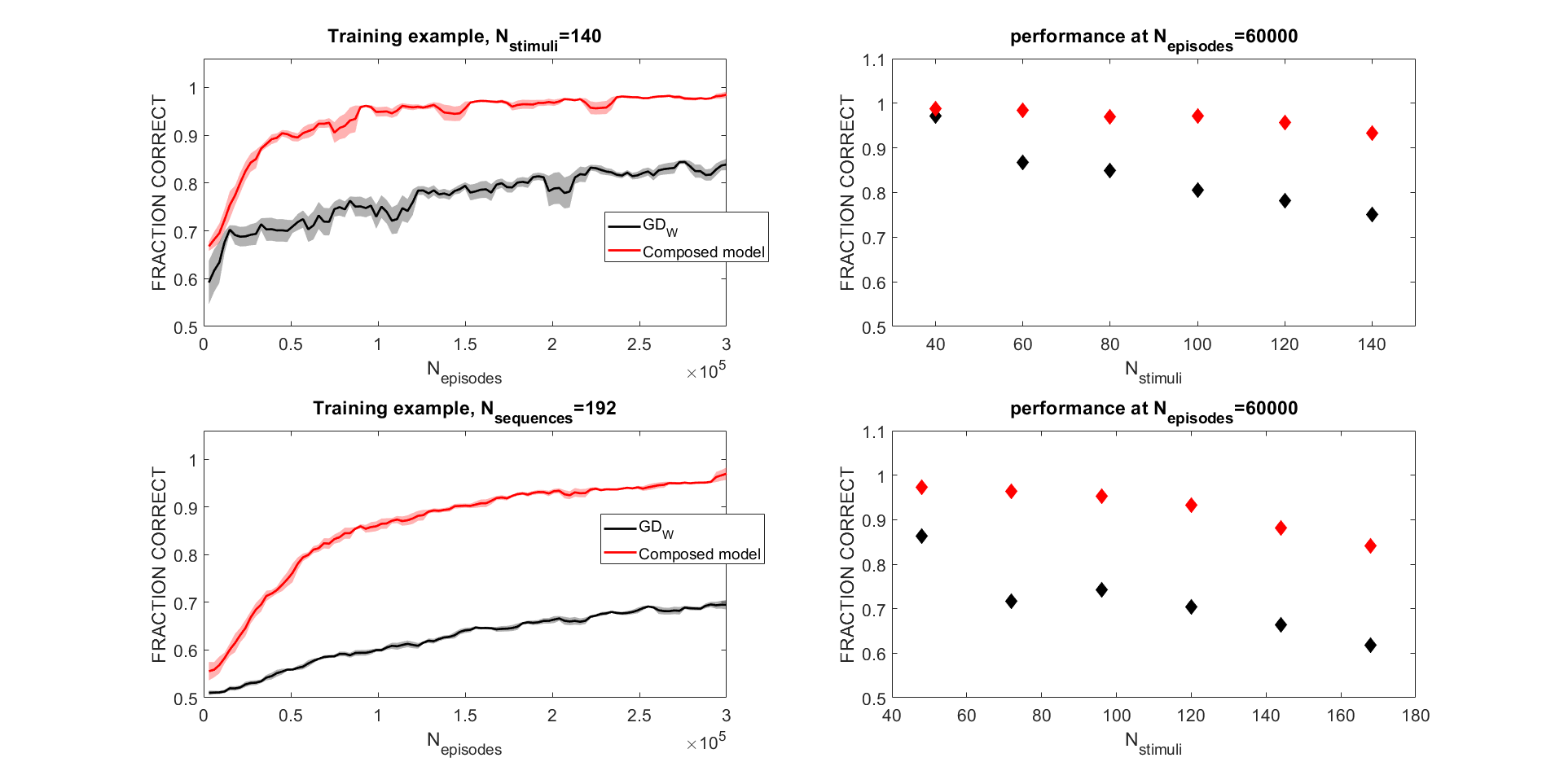}}
    \caption{\small{Performance of the proposed algorithm in a Reinforcement Learning framework. }}
    \label{Figure7}
\end{figure*}

\section{Reinforcement Learning}

In this section the learning on the thresholds is performed through  Reinforcement Learning to test the generalization ability of the proposed models. In this scenario the agent has to make a choice in order to receive a feedback from the environment, which provides information related to the action chosen only. The two tasks faced are an  example of the N-bandits problem where the decision over the N possible actions is made after a temporal interval $\Delta t$ in which the reservoir integrates the input signal and evolves through time. 
The specific algorithm exploited is Q-learning with a softmax policy.
Adopting a RL formalism, the output $y$ of the neural network corresponds to the $Q$ value and  $y_{true}$ corresponds to the reward $R$.
The error function to be minimized is

\begin{align}
E_{RL}=\Big[ R-Q(a,\textbf{x}(t)) \Big]^2 \\
Q(a,t)=\sum_i W_{ai} relu\big[ V_i(t)-\theta_i \big]
\label{RL_E}
\end{align}

where the sum over the N possible outcomes of eq.\ref{E} is no longer present in the Reinforcement Learning rule. 
However, in some cases we will use a batch version of eq.\ref{RL_E} to improve the performance    
\begin{align}
E_{RL}=\sum_{k=1}^{N_{batch}}\Big[ R_k-Q(a_k,\textbf{x}_k(t)) \Big]^2 \\
\end{align}   
  
The qualitative arguments regarding the meaning of the application of a gradient descent algorithm on the thresholds, described in the section \textit{Gradient Descent} in \textit{Methods},  are valid also in this case thanks to the sampling over different actions. For clarity, the updating rule obtained by applying a gradient descent method to the vector $\bm{\theta}$ is 

\begin{equation}
\nonumber
\Delta \theta_i 
=-\eta \big[ R-Q(a,\textbf{x}(t)) \big] W^{out}_{ai}H\big(x_i(t)\big)
\end{equation}

The performance for the composite algorithm and $GD_{W}$ are reported in fig.\ref{Figure7}.

\section{Discussion}
The model proposed efficiently optimizes the values of the thresholds to reach a sparsity level that can increase the specificity of the nodes in the reservoir.
The parameter space of the thresholds is searched with a gradient descent algorithm and a stochastic perturbation of a global parameter. The mixture of the two approaches is necessary since the gradient descent on $\theta$ ($GD_{\theta}$) learns a suboptimal value of the average of the distribution of $\theta$, while the implemented stochastic search optimizes a single global parameter for the whole reservoir. It would be possible to adopt this random perturbation to each single neuron separately, but the algorithm would have to learn a high number of parameters through stochastic search, which would dramatically increase the convergence time, while the learning of a global threshold allows for a fast learning of the mean of the distribution.     
The improvements in the performance achievable through the application of the model comes with the price of learning the thresholds. Indeed, future research work should focus on understanding the dependence of the optimization processes on the three different variables ($\theta_i$,$\theta_g$ and $W^{out}$) involved in the learning procedure. Ideally,  hyperparameters should be defined (or learned) and interpreted as functions of the features of the external input. \newline
Since the model outperforms the non sparse network in all the tasks analysed, we consider it as a promising algorithm to induce sparse representations that increase the accuracy and convergence speed of reservoirs of neurons and, possibly, recurrent neural networks in general.

\nocite{lukovsevivcius2009reservoir} 
\nocite{bhandawat2007sensory}
\nocite{werfel2004learning}
\nocite{waddell2013reinforcement}
\nocite{huang2007sparse}
{\small
\bibliographystyle{unsrt}
\bibliography{Manneschi_Lin_Vasilaki}

\begin{thebibliography}{10}

\bibitem{rolls1995sparseness}
Edmund~T Rolls and Martin~J Tovee.
\newblock Sparseness of the neuronal representation of stimuli in the primate
  temporal visual cortex.
\newblock {\em Journal of neurophysiology}, 73(2):713--726, 1995.

\bibitem{bhandawat2007sensory}
Vikas Bhandawat, Shawn~R Olsen, Nathan~W Gouwens, Michelle~L Schlief, and
  Rachel~I Wilson.
\newblock Sensory processing in the drosophila antennal lobe increases
  reliability and separability of ensemble odor representations.
\newblock {\em Nature neuroscience}, 10(11):1474, 2007.

\bibitem{lin2014sparse}
Andrew~C Lin, Alexei~M Bygrave, Alix De~Calignon, Tzumin Lee, and Gero
  Miesenb{\"o}ck.
\newblock Sparse, decorrelated odor coding in the mushroom body enhances
  learned odor discrimination.
\newblock {\em Nature neuroscience}, 17(4):559, 2014.

\bibitem{huang2011learning}
Junzhou Huang, Tong Zhang, and Dimitris Metaxas.
\newblock Learning with structured sparsity.
\newblock {\em Journal of Machine Learning Research}, 12(Nov):3371--3412, 2011.

\bibitem{hastie2015statistical}
Trevor Hastie, Robert Tibshirani, and Martin Wainwright.
\newblock {\em Statistical learning with sparsity: the lasso and
  generalizations}.
\newblock Chapman and Hall/CRC, 2015.

\bibitem{candes2008enhancing}
Emmanuel~J Candes, Michael~B Wakin, and Stephen~P Boyd.
\newblock Enhancing sparsity by reweighted l 1 minimization.
\newblock {\em Journal of Fourier analysis and applications}, 14(5-6):877--905,
  2008.

\bibitem{NIPS2016_6504}
Wei Wen, Chunpeng Wu, Yandan Wang, Yiran Chen, and Hai Li.
\newblock Learning structured sparsity in deep neural networks.
\newblock In D.~D. Lee, M.~Sugiyama, U.~V. Luxburg, I.~Guyon, and R.~Garnett,
  editors, {\em Advances in Neural Information Processing Systems 29}, pages
  2074--2082. Curran Associates, Inc., 2016.

\bibitem{srivastava2014dropout}
Nitish Srivastava, Geoffrey Hinton, Alex Krizhevsky, Ilya Sutskever, and Ruslan
  Salakhutdinov.
\newblock Dropout: a simple way to prevent neural networks from overfitting.
\newblock {\em The journal of machine learning research}, 15(1):1929--1958,
  2014.

\bibitem{rasmussen2012model}
Peter~M Rasmussen, Lars~K Hansen, Kristoffer~H Madsen, Nathan~W Churchill, and
  Stephen~C Strother.
\newblock Model sparsity and brain pattern interpretation of classification
  models in neuroimaging.
\newblock {\em Pattern Recognition}, 45(6):2085--2100, 2012.

\bibitem{jaeger2007optimization}
Herbert Jaeger, Mantas Luko{\v{s}}evi{\v{c}}ius, Dan Popovici, and Udo Siewert.
\newblock Optimization and applications of echo state networks with
  leaky-integrator neurons.
\newblock {\em Neural networks}, 20(3):335--352, 2007.

\bibitem{takemura2017connectome}
Shin-ya Takemura, Yoshinori Aso, Toshihide Hige, Allan Wong, Zhiyuan Lu, C~Shan
  Xu, Patricia~K Rivlin, Harald Hess, Ting Zhao, Toufiq Parag, et~al.
\newblock A connectome of a learning and memory center in the adult drosophila
  brain.
\newblock {\em Elife}, 6:e26975, 2017.

\bibitem{liu2016gap}
Qingqing Liu, Xing Yang, Jingsong Tian, Zhongbao Gao, Meng Wang, Yan Li, and
  Aike Guo.
\newblock Gap junction networks in mushroom bodies participate in visual
  learning and memory in drosophila.
\newblock {\em Elife}, 5:e13238, 2016.

\bibitem{caron2013random}
Sophie~JC Caron, Vanessa Ruta, LF~Abbott, and Richard Axel.
\newblock Random convergence of olfactory inputs in the drosophila mushroom
  body.
\newblock {\em Nature}, 497(7447):113, 2013.

\bibitem{jaeger2001echo}
Herbert Jaeger.
\newblock The "echo state" approach to analysing and training recurrent neural
  networks-with an erratum note.
\newblock {\em Bonn, Germany: German National Research Center for Information
  Technology GMD Technical Report}, 148(34):13, 2001.

\bibitem{jaeger2002tutorial}
Herbert Jaeger.
\newblock {\em Tutorial on training recurrent neural networks, covering BPPT,
  RTRL, EKF and the" echo state network" approach}, volume~5.
\newblock GMD-Forschungszentrum Informationstechnik Bonn, 2002.

\bibitem{hallem2006coding}
Elissa~A Hallem and John~R Carlson.
\newblock Coding of odors by a receptor repertoire.
\newblock {\em Cell}, 125(1):143--160, 2006.

\bibitem{olsen2010divisive}
Shawn~R Olsen, Vikas Bhandawat, and Rachel~I Wilson.
\newblock Divisive normalization in olfactory population codes.
\newblock {\em Neuron}, 66(2):287--299, 2010.

\bibitem{luo2010generating}
Sean~X Luo, Richard Axel, and LF~Abbott.
\newblock Generating sparse and selective third-order responses in the
  olfactory system of the fly.
\newblock {\em Proceedings of the National Academy of Sciences},
  107(23):10713--10718, 2010.

\bibitem{parnas2013odor}
Moshe Parnas, Andrew~C Lin, Wolf Huetteroth, and Gero Miesenb{\"o}ck.
\newblock Odor discrimination in drosophila: from neural population codes to
  behavior.
\newblock {\em Neuron}, 79(5):932--944, 2013.

\bibitem{krishnamurthy2017disorder}
Kamesh Krishnamurthy, Ann~M Hermundstad, Thierry Mora, Aleksandra~M Walczak,
  and Vijay Balasubramanian.
\newblock Disorder and the neural representation of complex odors: smelling in
  the real world.
\newblock {\em arXiv preprint arXiv:1707.01962}, 2017.

\bibitem{kuczera1998monte}
George Kuczera and Eric Parent.
\newblock Monte carlo assessment of parameter uncertainty in conceptual
  catchment models: the metropolis algorithm.
\newblock {\em Journal of Hydrology}, 211(1-4):69--85, 1998.

\bibitem{lukovsevivcius2009reservoir}
Mantas Luko{\v{s}}evi{\v{c}}ius and Herbert Jaeger.
\newblock Reservoir computing approaches to recurrent neural network training.
\newblock {\em Computer Science Review}, 3(3):127--149, 2009.

\bibitem{werfel2004learning}
Justin Werfel, Xiaohui Xie, and H~Sebastian Seung.
\newblock Learning curves for stochastic gradient descent in linear feedforward
  networks.
\newblock In {\em Advances in neural information processing systems}, pages
  1197--1204, 2004.

\bibitem{waddell2013reinforcement}
Scott Waddell.
\newblock Reinforcement signalling in drosophila; dopamine does it all after
  all.
\newblock {\em Current opinion in neurobiology}, 23(3):324--329, 2013.

\bibitem{huang2007sparse}
Ke~Huang and Selin Aviyente.
\newblock Sparse representation for signal classification.
\newblock In {\em Advances in neural information processing systems}, pages
  609--616, 2007.

\end{thebibliography}
}

\end{document}